%% file: main.tex
\documentclass[11pt]{article}

\usepackage[margin=1in]{geometry}

\usepackage{amsmath}
\usepackage{amssymb}
\usepackage{amsthm}

\usepackage[utf8]{inputenc}
\usepackage[T1]{fontenc}
\usepackage{lmodern}
\usepackage{microtype}

\usepackage[colorlinks=true,linkcolor=blue,citecolor=blue,urlcolor=blue]{hyperref}
\usepackage{natbib}

\newtheorem{theorem}{Theorem}

\title{Gradient Descent as Implicit EM in Distance-Based Neural Models}
\author{Alan Oursland}
\date{December 2025; Revised July 2026}

\begin{document}

\maketitle

\input{abstract.tex}

\input{sec-introduction.tex}

\input{sec-geometric-substrate.tex}

\input{sec-main-result.tex}

\input{sec-three-regimes.tex}

\input{sec-prior-work.tex}

\input{sec-limits.tex}

\input{sec-discussion.tex}

\input{sec-conclusion.tex}

\bibliographystyle{plainnat}
\bibliography{references}

\end{document}

%% file: abstract.tex
\begin{abstract}

Neural networks trained with standard objectives exhibit behaviors characteristic of probabilistic inference: soft clustering, prototype specialization, and Bayesian uncertainty tracking. These phenomena appear across architectures---in attention mechanisms, classification heads, and energy-based models---yet existing explanations rely on loose analogies to mixture models or post-hoc architectural interpretation. We provide a direct explanation. For any objective with log-sum-exp structure over distances or energies, the gradient with respect to each distance is exactly the negative posterior responsibility of the corresponding component: $\partial L / \partial d_j = -r_j$. The identity is algebraic, requiring only differentiability; it is a specialization of Fisher's identity from the classical EM literature, and its significance here is its address: standard neural objectives instantiate it without modification. The immediate consequence is that gradient descent on such objectives performs a generalized expectation-maximization implicitly, with responsibilities arising as gradients to be applied rather than auxiliary variables to be computed. No explicit inference algorithm is required because inference is embedded in the optimization. This result unifies three regimes of learning under a single mechanism: unsupervised mixture modeling, where responsibilities are fully latent; attention, where responsibilities are conditioned on queries; and cross-entropy classification, where supervision clamps responsibilities to targets. Our claims live at training time: the responsibility-weighted gradient dynamics recently documented in transformers follow necessarily from the objective's geometry. The in-context Bayesian computation that trained transformers perform at inference time is the endpoint of these dynamics, not their per-step content. For this class of objectives, learning and inference are the same process viewed at different levels.

\end{abstract}

%% file: sec-introduction.tex
\section{Introduction}

Neural networks trained with standard objectives repeatedly exhibit behaviors associated with probabilistic inference: soft clustering, prototype specialization, uncertainty tracking, and mixture-model dynamics. These phenomena appear across architectures, including attention mechanisms, classification heads, and energy-based models, yet their origin has remained unclear. They are variously attributed to scale, to architectural particulars, or left unexplained. We argue that such behaviors are neither emergent nor accidental, but are necessary consequences of the geometry of common objective functions.

\subsection{The Puzzle}

A range of phenomena appear in trained neural networks without being designed in. Attention heads in transformers learn to specialize, each routing information for distinct semantic roles. Classification networks partition representation space into regions that behave like mixture components. Deep networks trained on noisy data exhibit robustness patterns reminiscent of Bayesian inference, down-weighting outliers and tracking uncertainty across inputs. These behaviors arise without explicit probabilistic modeling, mixture-model architectures, or any algorithm resembling expectation-maximization.

The standard explanations are unsatisfying. One view holds that these are emergent properties of scale, with sufficient parameters and data giving rise to statistical structure through unspecified mechanisms. Another treats them as architectural coincidences, artifacts of specific design choices such as softmax normalization or residual connections. A third offers loose analogies: attention is ``like'' soft clustering; cross-entropy ``resembles'' a mixture model. None of these accounts explains why these specific behaviors appear rather than others, or why they appear so reliably across different architectures and tasks.

\subsection{Recent Evidence}
\label{sec:scoping}

Recent work by \citet{agarwal2025geometry,agarwal2025gradient} has sharpened this puzzle. In controlled experimental settings---``Bayesian wind tunnels'' where the true posterior is analytically known---small transformers match the analytic posterior's entropy, position by position, to within $10^{-3}$--$10^{-4}$ bits: agreement at the resolution of the arithmetic. Capacity-matched MLPs trained under identical conditions fail catastrophically, and selective state-space models succeed on a subset of the tasks, suggesting that the phenomenon depends on content-based routing rather than on optimization alone.

The gradient dynamics reported by \citet{agarwal2025gradient} are equally striking. Attention weights tend to stabilize early in training while value vectors continue to refine, a two-timescale structure that mirrors the E-step and M-step of classical expectation-maximization. Values receive updates weighted by attention, precisely as prototypes receive updates weighted by responsibilities in mixture models. The authors provide a complete first-order analysis showing that this structure is systematic rather than incidental.

A scoping distinction is essential to everything that follows, as two different processes are in play. The \emph{Bayesian inference} demonstrated in the wind tunnels is in-context computation: a frozen, trained network tracking the posterior predictive over latent task variables as a sequence unfolds at inference time, a point made explicit in revised versions of both papers. The \emph{EM-like dynamics} are a training-time phenomenon: the gradient analysis, in the authors' words, ``lives at the EM/SGD level,'' explaining how cross-entropy training sculpts the geometry that the frozen network later uses. The claims of the present paper reside entirely at this second level. We analyze what gradient descent does during training, not what the learned function computes in context.

Within this training-time level, \citet{agarwal2025gradient} are careful about the status of the EM correspondence. Their gradient laws are exact consequences of cross-entropy, but they frame the EM correspondence itself as ``a mechanistic correspondence'' offered as ``an interpretive framework'': no surrogate objective with guaranteed monotonic improvement is derived. They likewise decline the variational reading of the dynamics: because value updates are driven by the backpropagated error signal rather than by observed data, values move to explain the error geometry rather than to maximize a likelihood over inputs---``the analogy is structural rather than variational.'' Their account thus leaves open the probabilistic status of the training dynamics: whether the responsibilities that gate learning are posteriors of anything at all.

\subsection{This Paper}

Before stating what this paper shows, it is worth stating where it came from, because the route explains the framing. The result was not found by interpreting a trained network. It came from taking the distance-based reading of neural outputs \citep{oursland2024interpreting}---in which a linear unit $y = w^\top x + b$ is read as a whitened Mahalanobis component $\lambda^{-1/2} v^\top (x - \mu)$ of a learned distribution---as a design constraint and asking the engineering question it forces: if a network's outputs are distances, what loss function should consume them? Following the types---exponentiate a distance to obtain a likelihood, normalize to compare alternatives, take the logarithm to return to the distance domain---leads directly to the log-sum-exp objective, a smooth ``distance to the nearest prototype'' (Section 2.2). Differentiating that objective to understand its training behavior is what exposed the identity below. The mixture model was not imposed on the loss; it was discovered inside it. The paper is organized to preserve that direction of travel: geometry first, probability derived.

This paper supplies the probabilistic status that the caveat of \citet{agarwal2025gradient} leaves open, and delimits it. We show that for objectives with log-sum-exp structure over distances or energies, the gradient of the loss with respect to each distance is exactly the posterior responsibility of the corresponding component. This is an algebraic identity rather than an approximation or an analogy:

\begin{equation}
\frac{\partial L}{\partial d_j} = -r_j
\end{equation}

The identity itself is classical---it is Fisher's identity from the EM literature, specialized to log-sum-exp objectives---and the precise relationship between gradient descent and EM has been characterized before \citep{xu1996convergence,neal1998view}. What has not been noted is that standard neural objectives instantiate this identity, unmodified, over the distance-based representations that ordinary networks compute. When the log-sum-exp is the loss, the EM correspondence is variational rather than structural: the responsibilities are posterior probabilities of an implicit mixture, and gradient descent performs expectation-maximization continuously rather than in discrete alternating steps. The forward pass computes unnormalized likelihoods, normalization yields responsibilities, and backpropagation delivers responsibility-weighted updates to parameters, without auxiliary latent variables or a separate inference algorithm. In this setting the gradient and the responsibility coincide. When the softmax instead sits inside the network, as in attention, the same calculus recovers the structural laws of \citet{agarwal2025gradient} and explains why the variational reading applies at the loss level and becomes structural in the interior (Section 4.2).

This reframes the relationship between optimization and inference. Under the objectives we analyze, inference is neither a separate algorithmic layer added on top of learning nor a post-hoc interpretation of learned representations: inference and optimization are the same computation viewed at different levels of abstraction. On this account, the Bayesian structure observed by \citet{agarwal2025gradient} is forced by the geometry of the loss rather than emergent.

\subsection{Contributions}

The contributions of this paper are deliberately narrow and can be summarized as three claims at increasing levels of abstraction.

\textbf{One theorem.} For any objective of the form $L = \log \sum_j \exp(-d_j)$, the gradient with respect to each distance satisfies $\partial L / \partial d_j = -r_j$, where $r_j$ is the normalized exponential---the responsibility of component $j$. The identity is a specialization of Fisher's identity from the classical EM literature (Section 5.2) and requires no conditions beyond differentiability. What is new is not the identity but its address: standard neural objectives, including cross-entropy, attention, and mixture likelihoods, instantiate it without modification.

\textbf{One interpretation.} This identity implies that gradient descent on distance-based log-sum-exp objectives performs implicit expectation-maximization. The E-step is the forward pass, the M-step is the parameter update, and responsibilities are never explicitly computed because they coincide with the gradients. Neural training does not approximate EM in this setting; it realizes it.

\textbf{One unification.} The same mechanism manifests across three regimes depending on the constraints imposed. In the \emph{unsupervised} regime, responsibilities are fully latent and prototypes compete freely. In the \emph{conditional} regime (attention), responsibilities are recomputed per query over a shared prototype family. In the \emph{constrained} regime (cross-entropy classification), supervision clamps one responsibility while competition persists among alternatives. These three regimes are one phenomenon under different boundary conditions.

%% file: sec-geometric-substrate.tex
\section{The Geometric Substrate}

The main result of this paper---that responsibilities emerge as gradients---rests on a specific geometric foundation. This section establishes that foundation. We first summarize the interpretation of neural network outputs as distances or energies rather than confidences, importing this result from prior work. We then define the class of log-sum-exp objectives to which our analysis applies. Finally, we recall the structure of classical expectation-maximization, which serves only as the reference point against which implicit EM will be compared.

\subsection{Distance-Based Representations}
\label{sec:distance}

The standard interpretation of neural network outputs treats them as confidences or scores indicating the strength of evidence for a hypothesis. A logit is high when the network ``believes'' in a class; an attention score is high when a query ``matches'' a key. This interpretation, while intuitive, obscures the geometric structure of the underlying computation.

In prior work \citep{oursland2024interpreting}, we developed a different interpretation: the outputs of standard neural networks are better understood as distances or energies relative to learned prototypes. The correspondence is direct. The Mahalanobis distance of a point $x$ from a Gaussian component with mean $\mu$, measured along a principal direction $v$ with eigenvalue $\lambda$, is

\begin{equation}
d(x) = |\lambda^{-1/2} v^\top (x - \mu)| = |w^\top x + b|, \qquad w = \lambda^{-1/2} v, \quad b = -\lambda^{-1/2} v^\top \mu,
\end{equation}

which is a linear layer followed by an absolute value. Standard ReLU networks compute the same quantity via the identity $|z| = \mathrm{relu}(z) + \mathrm{relu}(-z)$, decomposing the signed deviation into two half-space detectors. Under this view, a low output indicates proximity to a prototype and a high output indicates deviation, and probabilities are derived quantities, arising only after exponentiation and normalization transform distances into relative likelihoods. Empirical companions to that work test the distance-metric hypothesis directly \citep{oursland2024distance,oursland2025learn}.

The Mahalanobis form is what makes the passage to probability canonical rather than chosen. A Mahalanobis distance is, by construction, the negative logarithm of a Gaussian density up to constants: $d^2 = (x-\mu)^\top \Sigma^{-1} (x-\mu) = -2\log \mathcal{N}(x;\mu,\Sigma) + \text{const}$. To interpret an output as a Mahalanobis distance is therefore already to interpret it as a log-domain probabilistic quantity. Exponentiation, in the sections that follow, inverts this logarithm and recovers the density the distance came from, rather than imposing a probability model on the network. This is the sense in which the distance reading is the load-bearing step of the framework: for a generic score, $\exp(-d)$ is one transformation among many; for a Mahalanobis distance, it is the unique inverse of the definition. Everything downstream---likelihoods, responsibilities, implicit EM---follows from this single semantic commitment.

This interpretation reflects the structure of the computation itself: the weights of a linear layer define a basis, the biases define offsets along that basis, and the activation measures deviation. The computation is unchanged in either reading; what differs is the semantics assigned to it. Throughout this paper, we adopt the distance-based interpretation and refer to neural outputs as energies or distances interchangeably. We emphasize, however, that the main result does not depend on this reading: the gradient identity of Section~\ref{sec:main-result} is purely algebraic and holds for arbitrary energies. The distance interpretation supplies the semantics under which that identity deserves the name inference; it motivates the framework rather than carrying the mathematics.

\subsection{The Log-Sum-Exp Objective}
\label{sec:lse}

Given a set of distances or energies $\{d_1, d_2, \ldots, d_K\}$ computed for an input $x$, we consider objectives of the form

\begin{equation}
L(x) = \log \sum_{j=1}^{K} \exp(-d_j(x)).
\end{equation}

This objective admits two readings, and the order in which they are taken matters to this paper.

The first reading is geometric and requires no probability at all. The negative of $L$ is the soft minimum of the distances: for $K$ components,

\[
\min_j d_j - \log K \;\leq\; -L \;\leq\; \min_j d_j,
\]

so minimizing $-L$ minimizes, up to an additive constant, the distance from the input to the nearest prototype. The LSE objective is thus the smooth form of a distance-to-a-set: the loss that requires the input to be close to at least one prototype, softened enough to be differentiable. Under this reading, the gradient concentration of Theorem~\ref{thm:main}---signal flowing to the closest component---is the smooth analogue of a minimum being attained at its argmin. This design-first route is how the framework was arrived at: starting from distance-based representations and asking what objective should consume a distance, one is led to exponentiation (distance to likelihood), normalization (comparison across alternatives), and the logarithm (back to the distance domain), at which point the mixture model is discovered already inside the loss rather than imposed on it.

The second reading is probabilistic: if $\exp(-d_j)$ is read as the unnormalized likelihood that component $j$ generated the input, then $L$ is the log marginal likelihood, the log-probability that at least one component generated it. Maximizing $L$ encourages the model to place at least one prototype close to each input. The two readings pick out the same objective; the geometric one explains why it is natural, the probabilistic one explains what its gradients mean.

We adopt the following notation throughout. Let $P_j = \exp(-d_j)$ denote the unnormalized likelihood of component $j$, and let $Z = \sum_k P_k = \sum_k \exp(-d_k)$ denote the partition function. Define the responsibility of component $j$ as

\begin{equation}
r_j = \frac{P_j}{Z} = \frac{\exp(-d_j)}{\sum_k \exp(-d_k)}.
\end{equation}

The responsibilities are non-negative and sum to one. They represent the posterior probability that component $j$ is responsible for the input, under a uniform prior over components.

This structure appears throughout deep learning under various names. In cross-entropy classification, the logits $z_j$ act as negative distances ($d_j = -z_j$), and the objective becomes $L = -z_y + \log \sum_k \exp(z_k)$, where $y$ is the label: the LSE term plus a clamping term $-z_y$ contributed by supervision, the constrained regime of Section~\ref{sec:constrained}. In attention mechanisms, the scores $s_{ij}$ act as negative distances between query $i$ and key $j$, and softmax normalization produces responsibilities that weight the value vectors. These are direct instances of the LSE objective.

\subsection{Classical EM}

Expectation-maximization is the classical algorithm for fitting mixture models with latent assignments. It alternates between two steps.

In the E-step, responsibilities are computed: given current parameters, each data point is softly assigned to each component according to relative likelihood,

\[
r_j^{(t)} = \frac{P(x \mid \theta_j)}{\sum_k P(x \mid \theta_k)}.
\]

The responsibilities sum to one and represent the posterior probability that component $j$ generated the observation. (We present the uniform-mixing-weight case, matching the uniform prior of Section~\ref{sec:lse}; general EM weights each likelihood by a mixing proportion $\pi_j$ and updates the $\pi_j$ alongside the component parameters.)

In the M-step, parameters are updated: each component is adjusted to better fit the points assigned to it, weighted by responsibility,

\[
\theta_j^{(t+1)} = \arg\max_{\theta_j} \sum_i r_{ij}^{(t)} \log P(x_i \mid \theta_j).
\]

For Gaussian mixtures, this reduces to computing responsibility-weighted means and covariances. The key property is that every data point influences every component, with the influence gated by how much responsibility that component holds for that point.

Classical EM is discrete and alternating: compute all responsibilities, then update all parameters, then repeat. The E-step and M-step are separate procedures with distinct computational roles. This separation is algorithmic rather than fundamental: \citet{neal1998view} showed that both steps are coordinate ascent on a single free-energy objective, a view whose consequences for gradient-based training we develop in what follows.

%% file: sec-main-result.tex
\section{Main Result: Responsibilities Are Gradients}
\label{sec:main-result}

We now state and derive the central result: for log-sum-exp objectives over distances, the gradient with respect to each distance is exactly the negative responsibility of the corresponding component. The identity requires no approximations and holds for any model that computes distances and optimizes an LSE objective via gradient descent. The identity itself is classical: it is Fisher's identity---the gradient of an incomplete-data log-likelihood equals the posterior-expected complete-data gradient \citep{cappe2009online,bishop2006pattern}---specialized to a uniform-prior mixture over exponentiated distances (Section~\ref{sec:fisher}). The contribution of this paper is not the derivative but the recognition of where it applies: standard neural objectives instantiate this structure over distance-based representations without modification, so every gradient step of ordinary training is a responsibility-weighted update.

\subsection{Derivation}

Let $d_j(x)$ denote the distance or energy associated with component $j$ for input $x$. Define unnormalized likelihoods $P_j = \exp(-d_j)$ and the partition function $Z = \sum_k \exp(-d_k)$. Consider the log-sum-exp objective
\begin{equation}
L = \log \sum_{j=1}^{K} \exp(-d_j) = \log Z.
\end{equation}
We compute the gradient of $L$ with respect to a single distance $d_i$:
\[
\frac{\partial L}{\partial d_i} = \frac{1}{Z} \cdot \frac{\partial}{\partial d_i} \sum_{j=1}^{K} \exp(-d_j).
\]
Only the $i$-th term in the sum depends on $d_i$:
\[
\frac{\partial L}{\partial d_i} = \frac{1}{Z} \cdot \frac{\partial}{\partial d_i} \exp(-d_i) = \frac{1}{Z} \cdot (-1) \cdot \exp(-d_i).
\]
Simplifying:
\[
\frac{\partial L}{\partial d_i} = -\frac{\exp(-d_i)}{\sum_k \exp(-d_k)}.
\]
The right-hand side is the negative of the normalized exponential: the responsibility $r_i$ defined in Section~\ref{sec:lse}.

\begin{theorem}
\label{thm:main}
Let $L = \log \sum_j \exp(-d_j)$ for differentiable distances $d_j$. Then the gradient with respect to the $i$-th distance is the negative responsibility of component $i$:
\begin{equation}
\label{eq:main-identity}
\frac{\partial L}{\partial d_i} = -r_i, \qquad r_i = \frac{\exp(-d_i)}{\sum_k \exp(-d_k)}.
\end{equation}
\end{theorem}

\subsection{What This Means}

The theorem admits a direct interpretation. In classical EM, responsibilities are computed in the E-step and stored for use in the M-step. Under gradient descent on an LSE objective, the same quantities appear as the gradients themselves: any system that computes distances and optimizes this objective produces responsibilities in the course of backpropagation, where they act directly as the learning signal.

The identity is purely algebraic and holds regardless of interpretation; the EM reading additionally requires interpreting the $d_j$ as distances and the $r_j$ as posterior responsibilities, per Section~\ref{sec:distance}.

Consider what happens during training. The forward pass computes distances $d_j$ and evaluates the objective $L = \log \sum_j \exp(-d_j)$; the responsibilities are not computed as such. Backpropagation then computes $\partial L / \partial d_j = -r_j$, and the chain rule propagates this gradient to the parameters that produced $d_j$. The parameters of component $j$ thus receive gradient signal proportional to $r_j$, the responsibility of that component for the input. This is the M-step of expectation-maximization, performed within the gradient computation: the forward pass plays the role of the E-step, the backward pass the role of the M-step, fused rather than alternating.

The consequence is that gradient descent on LSE objectives performs a generalized EM implicitly. The precise relationship is inherited from classical results: the gradient is the responsibility-weighted update direction (Fisher's identity), and the EM step is that same gradient under a positive-definite preconditioner \citep{xu1996convergence}---the two methods share update structure, though not trajectories \citep{zhang2020comparing}. The discrete alternation of classical EM---compute responsibilities, then update parameters, then repeat---is likewise not fundamental: E and M are coordinate ascent on a single objective \citep{neal1998view}, and under gradient descent they collapse into continuous, simultaneous optimization. Every gradient step is a responsibility-weighted update; every network trained on these objectives has been performing responsibility-weighted learning throughout.

Throughout this paper, ``implicit EM'' refers to the emergence of responsibility-weighted parameter updates under gradient descent on log-sum-exp objectives. The term carries no claim of coordinate-ascent structure or convergence guarantees.

\subsection{Conditions}
\label{sec:conditions}

The result depends on three structural conditions. When all three are satisfied, implicit EM dynamics follow necessarily.

\textbf{Exponentiation.} The distances must be passed through an exponential to form unnormalized likelihoods $P_j = \exp(-d_j)$. This transformation converts additive differences in distance into multiplicative ratios in likelihood. Without exponentiation, differences in distance enter the gradient linearly rather than as likelihood ratios, and no responsibility structure can form. Under the Mahalanobis reading of Section~\ref{sec:distance}, this condition comes without additional assumptions: the distance is a negative log-density, and exponentiation merely inverts the logarithm.

\textbf{Normalization.} The likelihoods must be normalized across alternatives, either explicitly through softmax or implicitly through the log-sum-exp structure. Normalization induces competition: increasing the likelihood of one component necessarily decreases the relative likelihood of others. This competition makes responsibilities sum to one and distributes gradient signal according to relative fit. Without normalization, components operate independently, and there is no assignment structure.

\textbf{Gradient-based optimization.} The parameters must be updated via gradients of the objective. The identity $\partial L / \partial d_j = -r_j$ is a fact about the loss surface; it becomes a fact about learning dynamics only when gradient descent (or a variant) is used. Other optimization methods---evolutionary strategies, random search, discrete updates---would not automatically inherit the responsibility-weighted structure.

When these conditions hold, no additional design choice enables or disables the behavior: any architecture that computes distances, normalizes via exponentiation, and trains with gradients will exhibit implicit EM.

\subsection{Completing the Gaussian: The Volume Term}
\label{sec:volume}

The analysis so far treats the distances $d_j$ as given, which corresponds to mixture components of fixed shape. A full Gaussian likelihood carries an additional normalization term, the volume of each component's covariance, and neural objectives that omit it are exposed to metric collapse (Section~\ref{sec:collapse}). For the common case where distances are computed by linear layers, this missing term is itself expressible in network primitives, and Theorem~\ref{thm:main} extends to it verbatim.

Let component $j$ compute $z_j = W_j x + b_j$ with $W_j$ square and full-rank, and define the volume-corrected energy
\begin{equation}
\tilde{e}_j \;=\; \tfrac{1}{2}\|z_j\|^2 - \log\lvert\det W_j\rvert.
\end{equation}
Then $\exp(-\tilde{e}_j) = \lvert\det W_j\rvert \exp(-\tfrac{1}{2}\|W_j x + b_j\|^2)$, which is, up to the constant $(2\pi)^{-D/2}$ shared by all components, exactly the density $\mathcal{N}(x;\, \mu_j, \Sigma_j)$ with $\mu_j = -W_j^{-1} b_j$ and $\Sigma_j^{-1} = W_j^\top W_j$. The identity $\lvert\Sigma_j\rvert^{-1/2} = \lvert\det W_j\rvert$ is the change-of-variables Jacobian familiar from normalizing flows \citep{rezende2015variational}. The log-sum-exp objective over volume-corrected energies, $L = \log \sum_j \exp(-\tilde{e}_j)$, is therefore the exact log marginal likelihood of a full Gaussian mixture, with no proportionality constant and no fixed-covariance assumption, and Theorem~\ref{thm:main} applies unchanged: $\partial L / \partial \tilde{e}_j = -r_j$, where $r_j$ is now the exact posterior responsibility under fully normalized components.

Two consequences follow. First, implicit EM extends from means to covariances. The gradient of the objective with respect to the weight matrix is, per sample,
\begin{equation}
\frac{\partial L}{\partial W_j} \;=\; r_j \left( W_j^{-\top} - z_j x^\top \right),
\end{equation}
which is the responsibility times the gradient of the complete-data Gaussian log-likelihood: Fisher's identity at the parameter level. Gradient descent on this objective performs responsibility-weighted updates to the full component geometry, as the M-step of full-covariance EM does.

Second, the volume term removes the trivial collapse mode. Without it, the objective $\exp(-d_j)$ alone is maximized by shrinking the learned metric: driving $W_j \to 0$ maps every input to the prototype, sending all distances to zero and all likelihoods to one. The $\log\lvert\det W_j\rvert$ term sends that configuration to $-\infty$, and equilibrium is reached where the learned metric matches the data's covariance: growing $W_j$ is rewarded by volume but punished quadratically by distance, and vice versa. The classical single-point singularity of Gaussian mixture MLE---a component locking onto one data point and shrinking its variance without bound---remains, as it does in classical EM \citep{bishop2006pattern}, and is mitigated in practice by batch averaging or bounded parameterizations. Computationally, the determinant is cheap under standard parameterizations: for diagonal or triangular $W_j$, $\log\lvert\det W_j\rvert$ is a sum of log-diagonals, at $O(D)$ cost.

The framework is thus constructive as well as diagnostic: the same primitives that compute distances can compute the volume correction, the corrected objective is the exact mixture likelihood, and the implicit EM identity survives intact. Where Section~\ref{sec:collapse} explains why omitting the term invites collapse, this section shows what including it costs---one log-determinant---and what it buys. Whether practitioners should add the term in a given setting is a separate engineering question.

%% file: sec-three-regimes.tex
\section{Three Regimes of Implicit Inference}
\label{sec:regimes}

The same mechanism manifests differently under different constraints:

\subsection{Unsupervised Regime: Mixture Learning}

In the simplest case, the log-sum-exp objective operates without supervision. A model computes distances $d_j(x)$ from an input to each of $K$ components, and training minimizes
\begin{equation}
L = -\log \sum_{j=1}^{K} \exp(-d_j),
\end{equation}
the negative log marginal likelihood used in classical mixture model fitting. All components compete for every input, with no labels constraining the assignment.

By Theorem~\ref{thm:main}, the gradient of this minimization objective with respect to each distance is the responsibility itself: $\partial L / \partial d_j = +r_j$. (The sign flip relative to Section~\ref{sec:main-result} reflects the switch from maximizing the log-likelihood to minimizing its negative; the dynamics are identical.) A gradient step therefore decreases each distance in proportion to its responsibility. Components close to an input (low $d_j$, high $r_j$) receive strong gradient signal to move closer, while distant components (high $d_j$, low $r_j$) receive weak signal and remain largely unchanged. This is the M-step dynamic of Gaussian mixture models: prototypes update toward data points in proportion to their responsibility for those points.

The result is spontaneous specialization. Even from random initialization, components differentiate over training: each prototype drifts toward a region of input space for which it consistently takes high responsibility, while ceding other regions to competitors. Clustering emerges without being specified anywhere in the architecture.

This regime optimizes the same objective as classical EM on mixture models, with the same responsibilities governing both the fixed points and the path toward them.

\subsection{Conditional Regime: Attention Mechanisms}
\label{sec:attention}

Attention mechanisms instantiate the same structure in a conditional setting. For a query $q_i$ and a set of keys $\{k_j\}$, attention scores are computed as
\[
s_{ij} = \frac{q_i^\top k_j}{\sqrt{d_k}}.
\]
These scores act as negative distances: expanding $\|q_i - k_j\|^2 = \|q_i\|^2 - 2 q_i^\top k_j + \|k_j\|^2$, the dot product equals a negative squared distance up to norm terms, and the per-query term $\|q_i\|^2$ cancels under the softmax. Softmax normalization converts scores to attention weights,
\[
\alpha_{ij} = \frac{\exp(s_{ij})}{\sum_m \exp(s_{im})}.
\]
The attention weights satisfy the definition of responsibilities: they are non-negative, sum to one across keys, and measure the degree to which each key ``explains'' the query. The output is a responsibility-weighted combination of values, $o_i = \sum_j \alpha_{ij} v_j$, the estimation step of a mixture model conditioned on the query.

A distinction is required here. Theorem~\ref{thm:main} applies when the log-sum-exp is the objective. In a transformer, the attention softmax sits inside the network; the loss is cross-entropy at the output. The gradient of the loss with respect to an attention score is therefore not $-\alpha_{ij}$. The exact identities are obtained by the chain rule. Let $g_i = \partial L / \partial o_i$ be the error backpropagated to the attention output, and define the alignment $a_{ij} = g_i^\top v_j$ with attention-weighted mean $\bar{a}_i = \sum_k \alpha_{ik} a_{ik}$. Then
\begin{equation}
\frac{\partial L}{\partial v_j} = \sum_i \alpha_{ij} \, g_i
\qquad \text{and} \qquad
\frac{\partial L}{\partial s_{ij}} = \alpha_{ij} \left( a_{ij} - \bar{a}_i \right).
\end{equation}
The first identity is exact and has the same strength as Theorem~\ref{thm:main}: each value receives the downstream error scaled by its responsibility. This is the M-step for internal softmaxes, holding verbatim at every gradient step for any loss. The second identity shows that scores follow an advantage rule: a score rises only if its value is more useful to the downstream error than the current attention-weighted average. \citet{agarwal2025gradient} derive these two laws for transformer attention; they hold for any softmax module. Responsibilities still gate all learning, since both gradients carry the factor $\alpha_{ij}$, but where the loss-level LSE competes on proximity, the internal LSE competes on usefulness. Two consequences follow. Score gradients sum to zero across keys, so internal competition redistributes and cannot uniformly favor all components. And score updates vanish whenever the alignments equalize under the attention weights ($a_{ij} \approx \bar{a}_i$), of which full concentration ($\alpha_{ij^*} \to 1$) is the sharpest case, while value updates persist as long as error remains: the two-timescale dynamic observed by \citet{agarwal2025gradient} follows from the gradient forms.

There is also a second, exact sense in which attention computes an LSE gradient, in the forward pass itself. The gradient of $\log \sum_j \exp(q^\top k_j)$ with respect to the query is $\sum_j \alpha_j k_j$: the attention read-out is the gradient of a log-sum-exp energy, the retrieval dynamics of modern Hopfield networks \citep{ramsauer2020hopfield}. The forward pass takes an LSE gradient with respect to the state (inference); the backward pass takes responsibility-gated gradients with respect to the parameters (learning).

A second clarification concerns the prototypes. In a Gaussian mixture model, the component means are persistent parameters. In attention, the value vectors $v_j = W_V x_j$ are input-derived and transient. The persistent object is the value projection: by the chain rule, $\partial L / \partial W_V = \sum_{i,j} \alpha_{ij} \, g_i x_j^\top$, so the projection accumulates responsibility-weighted outer products. This is the M-step, applied not to fixed prototypes but to the parameters of a prototype-generating function. The EM dynamics thus operate across training rather than within a single forward pass, consistent with the scoping of Section~\ref{sec:scoping}. Over the course of training, $W_V$ specializes, learning to produce values suited to the queries that attend to them.

Attention is conditional mixture inference. The query specifies the context; the keys define candidate components; the values represent what each component contributes. The mixture is re-instantiated for every query, but the underlying parameters---$W_Q$, $W_K$, $W_V$---accumulate responsibility-weighted updates across all queries and all training examples. This is implicit EM in function space rather than data space.

\subsection{Constrained Regime: Cross-Entropy Classification}
\label{sec:constrained}

Cross-entropy classification introduces supervision. Supervision constrains the implicit EM structure but does not remove it.

In standard classification, the model produces logits $z_j$ for each class. Interpreting logits as negative distances ($d_j = -z_j$), the cross-entropy loss for a sample with label $y$ is
\[
L = -z_y + \log \sum_k \exp(z_k) = d_y + \log \sum_k \exp(-d_k).
\]
This is the log-sum-exp objective plus a clamping term that penalizes the distance to the correct class. Computing the gradient with respect to $d_j$,
\[
\frac{\partial L}{\partial d_j} = \mathbb{1}[j = y] - \frac{\exp(-d_j)}{\sum_k \exp(-d_k)} = \mathbb{1}[j = y] - r_j.
\]
(In logit form this is the familiar $\partial L / \partial z_j = p_j - \mathbb{1}[j=y]$; the sign flips because $d_j = -z_j$.) For incorrect classes ($j \neq y$), the gradient is $-r_j$: a descent step increases their distance, pushing each away in proportion to the responsibility it currently claims. For the correct class ($j = y$), the gradient is $1 - r_y \geq 0$: a descent step decreases its distance, with strength proportional to the responsibility it is missing.

The label acts as a clamp. In unsupervised mixture learning, responsibilities are entirely latent, determined only by the data. In cross-entropy, the label declares that the correct class should have responsibility 1, and the gradient corrects any deviation from this target. If the correct class already dominates ($r_y \approx 1$), the gradient is small. If incorrect classes hold probability mass ($r_y < 1$), the gradient is large.

Competition among incorrect classes remains intact. When the model misclassifies, the responsibility mass is distributed among wrong answers, and each receives gradient signal proportional to its share: the classes with highest $r_j$ are penalized most strongly. The repulsion is responsibility-weighted rather than uniform.

The clamping term directs the EM dynamics rather than replacing them. The M-step still updates components in proportion to their responsibilities, but supervision biases the process toward the clamped assignment. This may explain why cross-entropy is so effective despite its simplicity: it inherits the soft competition and automatic weighting of mixture models while channeling those dynamics toward a supervised target.

\subsection{The Taxonomy}

The three regimes---unsupervised, conditional, constrained---differ in what is observed and what is latent. They also differ in where the log-sum-exp sits. When the LSE is the loss (mixture likelihood, cross-entropy), gradients equal responsibilities, signed by supervision. When the LSE is inside the network (attention), value gradients are exactly responsibility-weighted while score gradients are responsibility-gated advantages. In every case the regimes share a common structure, exponentiation of distances followed by normalization across alternatives: that structure produces responsibilities, responsibilities produce implicit EM, and what varies is which quantity the responsibilities weight.

The critical factor is normalization. When outputs are normalized---whether by softmax, by the log-sum-exp partition function, or by any operation that forces a sum-to-one constraint---components compete: an increase in one component's likelihood necessarily decreases the relative likelihood of others. Through this competition, each input is probabilistically allocated across components, and gradients distribute accordingly.

Removing normalization collapses the structure. Consider objectives based on Gaussian kernels without a partition function, such as maximum correntropy over $K$ prototypes:
\[
L = \sum_{j=1}^{K} \left( 1 - \exp\!\left( -\frac{d_j^2}{\sigma^2} \right) \right).
\]
Up to a constant, this is the LSE objective's inner sum with the logarithm removed, and the logarithm is what creates the partition-function coupling. Here each component operates independently. A point far from all prototypes produces vanishing gradients for all of them---not low responsibility, but the absence of any responsibility structure, since no competition exists to distribute the assignment. There is no implicit E-step because there are no responsibilities. The objective gains robustness to outliers, since points far from all prototypes are effectively ignored, but loses the assignment structure entirely.

This clarifies the design space. Exponentiation converts distances to likelihoods; normalization converts likelihoods to responsibilities. With both, implicit EM follows necessarily; with exponentiation alone, the model gains robustness but forgoes the assignment structure. The choice of objective is a choice about whether the model assigns or ignores, and it is made at the level of the loss function rather than the architecture.

%% file: sec-prior-work.tex
\section{Relation to Prior Work}

The result derived here depends on prior work for its geometric foundation and gains significance from recent empirical findings that it explains. This section positions the contribution relative to four bodies of work: the distance-based interpretation of neural representations, classical results connecting EM to gradient methods, recent evidence of Bayesian structure in transformers, and energy-based learning.

\subsection{Prior Work on Distance-Based Representations}

In prior work \citep{oursland2024interpreting}, we established the geometric interpretation on which this paper depends. That work shows that standard neural network layers---affine transformations followed by ReLU or absolute value activations---compute quantities that behave as distances from learned reference structures. Outputs are read as deviations rather than confidences, measuring how far an input lies from a reference structure defined by the weights. The interpretation reflects the algebraic form of the operations rather than an assumption imposed on them (Section~\ref{sec:distance}).

The present work takes this geometric substrate as given and asks a different question: what happens when distance-based representations are optimized under standard objectives? The prior work characterizes what networks represent and is silent on how they learn. The contribution here is to show that log-sum-exp objectives over distances produce responsibility-weighted gradients, and that this induces implicit EM.

The two results are complementary. The prior work establishes that neural networks compute distances; the present work shows that optimizing distances under LSE objectives performs inference. Neither subsumes the other. Without the distance interpretation, the identification of gradients with responsibilities is a formal curiosity; without the gradient identity, the distance interpretation describes static representations with no account of how they emerge.

\subsection{Classical Results Connecting EM and Gradient Methods}
\label{sec:fisher}

The identity at the center of this paper is not new to statistics. It is a specialization of Fisher's identity: for any latent-variable model, the gradient of the incomplete-data log-likelihood equals the posterior expectation of the complete-data gradient \citep[see, e.g.,][]{cappe2009online}. For a mixture model, that posterior is exactly the vector of responsibilities, so marginal-likelihood gradients are responsibility-weighted by construction; the Gaussian-mixture case is textbook material \citep[\S 9.2]{bishop2006pattern}. Theorem~\ref{thm:main} is Fisher's identity for a uniform-prior mixture with component likelihoods $\exp(-d_j)$, differentiated with respect to the distances themselves.

The relationship between gradient descent and EM is also precisely characterized. \citet{xu1996convergence} prove that the EM update for Gaussian mixtures is a preconditioned gradient step on the log-likelihood, with an explicit positive-definite projection matrix; \citet{salakhutdinov2003optimization} use the expected-gradient identity to build direct optimizers for latent-variable models and characterize when EM behaves like a quasi-Newton method. \citet{neal1998view} show that the E-step and M-step are coordinate ascent on a single free-energy objective, so the discrete alternation of classical EM is an algorithmic choice rather than a structural fact. Our claim that inference and optimization fuse under LSE objectives inherits its precise form from these results: the instantaneous gradient is the responsibility-weighted update direction (Fisher), and the EM step is that same gradient under a positive-definite rescaling \citep{xu1996convergence}.

At the architectural level, the EM/attention connection also has a lineage. Modern Hopfield networks \citep{ramsauer2020hopfield} identify transformer attention with retrieval dynamics on a log-sum-exp energy: the attention output is the gradient of an LSE with respect to the query. Neural Expectation Maximization \citep{greff2017neural}, EM attention modules \citep{li2019expectation}, and EM capsule routing \citep{hinton2018matrix} all build explicit EM iterations into networks.

Given this literature, the contribution of the present paper is deliberately not the identity itself. It is the recognition that standard neural objectives---softmax cross-entropy, attention, mixture likelihoods---already instantiate Fisher's identity over distance-based representations without modification, and the analysis of how one mechanism manifests across the unsupervised, conditional, and constrained regimes. The classical results concern generative latent-variable models optimized deliberately; we argue the same structure operates, unnoticed, inside ordinary discriminative training.

\subsection{Agarwal et al.\ (2025a,b)}

\citet{agarwal2025geometry} provide striking empirical evidence that transformers implement Bayesian inference. In controlled settings where the true posterior is analytically known---their ``Bayesian wind tunnels''---small transformers match the analytic posterior's entropy, position by position, to within $10^{-3}$--$10^{-4}$ bits, while capacity-matched MLPs fail catastrophically. They also prove the endpoint statement: the population optimum of cross-entropy is the Bayes posterior predictive. That theorem is architecture-agnostic---it identifies what the optimum is, not how gradient training reaches it or what the intermediate quantities mean along the way.

\citet{agarwal2025gradient} move from statics to dynamics, analyzing the gradients of attention under cross-entropy training. They derive an advantage-based routing law for attention scores and a responsibility-weighted update for value vectors---the same identities we state in Section 4.2. They observe a two-timescale structure: attention patterns stabilize early while values continue to refine---mirroring the E-step and M-step of classical EM. The analysis is thorough and the EM parallel is explicit.

The authors are careful, however, about the status of the EM correspondence. Their gradient laws are derived, not analogized---the mechanics are exact consequences of cross-entropy. They present the attention weights as posterior responsibilities of a latent routing variable, but decline the variational reading of the dynamics: because the value updates are driven by the backpropagated error signal rather than by observed data, no likelihood over values is optimized, and they explicitly disclaim a surrogate objective with monotonic-improvement guarantees. In their framing, the EM connection is a mechanistic correspondence, not the optimization of any latent-variable likelihood.

The present work supplies the probabilistic status that this caveat leaves open, and delimits it. When the log-sum-exp is the loss (a mixture likelihood, or cross-entropy at the output layer), the correspondence is variational: the gradient with respect to each distance is the posterior responsibility of an implicit mixture (Fisher's identity), and implicit EM holds as an identity. When the softmax sits inside the network, Section~\ref{sec:attention} recovers their two laws and gives their caveat precise form: internal competition is over downstream usefulness rather than likelihood, which is why the variational reading stops at the loss level. The generality of the internal-softmax identities bears on their conjectured framework for content-based routing: the advantage structure they conjecture for arbitrary routing mechanisms holds, by the derivation of Section~\ref{sec:attention}, for any softmax-normalized routing regardless of the surrounding architecture, leaving the selective state-space case they identify as the open instance. Their population-optimum theorem is the endpoint statement of the loss-level case; the responsibility-gradient identity is its per-step counterpart.

A boundary must also be drawn. The authors are explicit that the Bayesian inference their experiments demonstrate is in-context computation at inference time---the posterior predictive over latent task variables, tracked by frozen weights as a sequence unfolds---and a third paper extends the geometric findings to production language models \citep{agarwal2025scaling}. Our framework does not explain that in-context computation. The responsibility-gradient identity is a statement about parameter updates during training; it connects to the inference-time geometry only through the endpoint theorem---training under cross-entropy drives the network toward the Bayes posterior predictive---and through their Paper II's demonstration that the training dynamics build the geometric machinery. Explaining how a frozen transformer performs sequential belief updating in context is their contribution, not ours.

The relationship between the contributions is therefore complementary and level-separated. Agarwal et al.\ establish the empirical geometry, the in-context inference it supports, and the exact mechanics of the training gradients. We establish what those training gradients are probabilistically, and the conditions under which the EM reading is an identity rather than a correspondence.

\subsection{Other Connections}

The energy-based learning framework of \citet{lecun2006tutorial} provides the conceptual scaffolding for our formulation. That work reframes learning as the minimization of an energy function, with probabilities derived via exponentiation and normalization. The log-sum-exp objective analyzed here is a special case of the free-energy formulation in energy-based models, and the responsibility-gradient connection is the corresponding instance of Fisher's identity discussed above. What we take from the energy-based view is the ordering of primitives---energies first, probabilities derived---which is the same ordering our distance-based reading assigns to neural outputs.

\citet{dempster1977maximum} introduced expectation-maximization as an algorithm for maximum likelihood estimation with latent variables, with the E-step and M-step defined as discrete, alternating operations. The present work shows that for distance-based LSE objectives, the same update structure is realized by gradient descent: the forward pass computes responsibilities implicitly, and backpropagation applies them. This is consistent with the classical formulation---the discrete alternation is one realization of responsibility-weighted updating, and continuous gradient descent is another \citep[cf.][]{neal1998view}. In this setting, EM-structured updating is a property of the objective geometry rather than an algorithmic choice.

\citet{vaswani2017attention} introduced the transformer architecture with attention as its core mechanism, presented as a soft retrieval operation: queries attend to keys to aggregate values. The implicit EM perspective reframes attention as conditional mixture inference, with attention weights as responsibilities and value projections as prototype parameters. The interpretation is consistent with the original formulation and supplies a probabilistic semantics that the retrieval description does not aim at.

Mixture of Experts models \citep{jacobs1991adaptive} use explicit gating networks to route inputs to specialized subnetworks. The gating weights are responsibilities by another name. The difference is architectural: in Mixture of Experts, gating is a separate learned function; in standard attention and classification, responsibilities emerge as gradients of the objective without dedicated gating machinery. On the implicit EM view, explicit gating is not required to obtain responsibility-weighted updates, which any log-sum-exp objective produces automatically; what dedicated gating adds is architectural, separating expert parameters and enabling sparse, conditional computation.

%% file: sec-limits.tex
\section{Limits and Failure Modes}

The implicit EM framework applies under specific conditions, and when those conditions fail, so does the analysis. This section delineates the boundaries: when implicit EM does not arise, what pathologies can occur even when it does, and what phenomena lie outside its scope.

\subsection{When Implicit EM Does Not Arise}

The identity $\partial L / \partial d_j = -r_j$ depends on the log-sum-exp structure; removing normalization removes the identity.

Consider multi-label classification with independent sigmoids. Each class $j$ has its own output $\sigma(z_j)$, and the loss decomposes as a sum of independent binary cross-entropies, with no partition function coupling the classes. The gradient with respect to $z_j$ depends only on the target for class $j$, not on the outputs of other classes.

In this setting, responsibilities do not exist: no quantity sums to one across classes, and no soft assignment distributes an input among alternatives. Each output channel operates in isolation. A point can be equally close to all prototypes or far from all of them, and the gradients reflect independent errors rather than a redistribution.

This is an absence of the required objective structure rather than a failure of the architecture. Implicit EM arises from competition, and competition arises from normalization. Systems with independent outputs may learn useful representations, but they do not perform mixture inference and do not exhibit responsibility-weighted dynamics. The explanatory scope of the framework ends where normalization ends.

\subsection{Scale and Collapse}
\label{sec:collapse}

A full Gaussian mixture model includes a log-determinant term in the likelihood, a penalty on the volume of each component's covariance. This term prevents collapse: without it, a component could shrink its covariance to zero, placing infinite density on a single point and achieving unbounded likelihood. The log-determinant diverges as the covariance collapses, counterbalancing the density gain.

Most neural objectives omit this term. Cross-entropy and attention softmax operate on distances or scores without explicit volume penalties. The implicit EM dynamics still hold (gradients remain responsibility-weighted), but nothing in the objective prevents the learned metric from degenerating. A network can learn to map all inputs to nearby points, collapsing the distance structure and trivializing the responsibilities. (Section~\ref{sec:volume} shows that for linear-layer distances the exact volume term is available in network primitives at the cost of one log-determinant; the analysis here concerns the common case where it is omitted.)

In practice, collapse is often avoided through mechanisms external to the objective: weight decay regularizes the scale of projections, layer normalization constrains activation magnitudes, and architectural choices such as residual connections preserve signal diversity. These interventions stabilize the geometry, but none is derived from the objective whose pathology it controls.

The implicit EM framework clarifies why collapse is a risk. When components are updated in proportion to their responsibilities, a component that captures slightly more mass receives stronger gradients, captures more mass still, and can dominate entirely. This positive feedback is inherent to EM dynamics and is classically controlled by the volume term. Neural networks omit that control and rely on other mechanisms to fill the gap. The framework does not solve this problem; it explains why the problem exists.

\subsection{Supervision Constraints}

In the unsupervised regime, responsibilities are entirely latent: the data alone determines which components claim which inputs. Supervision changes this. Labels declare which component should take responsibility, overriding what the geometry would otherwise dictate.

This constraint is powerful but rigid. Cross-entropy training forces the correct class toward responsibility 1, regardless of whether the input lies near that class's prototype or far from all prototypes. An input equidistant from all class boundaries still receives a hard label; the model must assign it somewhere. The soft, graded structure of responsibilities persists among incorrect classes, but the correct class is clamped.

A consequence is the closed-world assumption. Softmax normalization guarantees that responsibilities sum to one, so some class must take full responsibility for every input. There is no ``none of the above'': an out-of-distribution input, no matter how anomalous, is assigned to whichever class has the lowest distance, and the model's confidence may be arbitrarily high.

This is a limitation of the objective rather than the mechanism. The implicit EM framework describes how responsibilities arise and how they weight learning; it does not claim that the resulting assignments are semantically correct or that the objective captures all desirable behaviors. Cross-entropy with softmax enforces assignment; if non-assignment is needed, the objective must change. Open-set recognition, out-of-distribution detection, and selective prediction require objectives that break the closed-world assumption, either by introducing an explicit reject option or by abandoning normalization entirely. The implicit EM analysis applies to whatever objective is chosen; it does not choose the objective.

\subsection{What This Framework Does Not Explain}

The implicit EM framework explains one phenomenon: the emergence of responsibility-weighted learning dynamics from distance-based objectives. Several major phenomena lie outside its scope.

In-context, inference-time Bayesian computation is not addressed. A trained transformer with frozen weights can track a posterior over latent task variables as a sequence unfolds \citep{agarwal2025geometry,agarwal2025scaling}; that is a property of the learned function, executed in activation space at inference time. Our identity concerns gradient updates to parameters during training. The two are connected---training under cross-entropy drives the network toward the Bayes posterior predictive, and the training dynamics build the geometry the frozen network uses---but the framework derives the per-step structure of learning, not the algorithm the learned function implements in context. The forward-pass log-sum-exp gradient of Section~\ref{sec:attention} describes a single attention read-out, not sequential belief updating across a context.

Generalization is not addressed. The framework describes the dynamics of training, not the inductive biases that enable performance beyond the training distribution. Scaling laws, the predictable relationships among model size, data, and performance, also lie outside the analysis; nothing in the gradient identity suggests how performance should scale with parameters or compute.

Long-horizon reasoning, planning, and sequential decision-making involve temporal structure that the framework does not capture. Implicit EM describes how a single input is softly assigned to components and how those components update. It does not describe how representations compose over time, how goals propagate backward through action sequences, or how models learn to search.

Emergent capabilities---the sudden appearance of qualitatively new behaviors at scale---remain unexplained. The framework offers no account of why certain abilities appear discontinuously or at specific thresholds of model size. If emergent capabilities arise from implicit EM dynamics, the connection is not obvious; if they arise from other mechanisms, the framework is silent.

These are boundaries of scope rather than failures of the analysis. The contribution is to identify and derive one mechanism precisely, leaving phenomena beyond it to complementary explanations.

%% file: sec-discussion.tex
\section{Discussion}

This section reflects on what the implicit EM framework unifies, what it implies about standard training, and what questions remain open.

\subsection{Unification}

The framework shows that Gaussian mixture models, attention mechanisms, and cross-entropy classification are one mechanism operating under different constraints rather than three methods with superficial similarities. In GMMs, responsibilities are fully latent. In attention, they are conditioned on queries and recomputed per input. In cross-entropy, they are partially clamped by supervision. The underlying dynamics---exponentiation, normalization, responsibility-weighted updates---are identical.

This suggests a shift in how neural network training is interpreted. Probability is often treated as primitive: distributions are defined, likelihoods derived, and objectives optimized. The implicit EM perspective inverts this order. Networks compute geometric quantities, deviations from learned structures, and probabilities arise only after exponentiation and normalization. Geometry precedes probability; inference is a consequence of optimization on geometric objectives.

Loss functions, in this view, are geometric priors rather than arbitrary choices tuned for performance. Cross-entropy encodes the assumption that inputs belong to discrete categories, one of which takes full responsibility. Log-sum-exp over distances assumes inputs arise from a mixture of latent causes. Correntropy assumes outliers should be ignored. Each objective induces a different geometry of assignment and a different pattern of gradient flow. Choosing a loss function is choosing a theory of how data relates to structure.

\subsection{Implications}

For interpretability, the framework offers a direct path from training dynamics to semantic structure. If responsibilities are gradients, then the assignments a network makes are present in the backward pass, computed at every training step, rather than hidden quantities requiring probes or post-hoc analysis. The question of which component is responsible for an input has an answer in the gradient itself. This does not solve interpretability---understanding why a component takes responsibility requires further analysis---but it locates the assignment structure in a quantity that is already computed.

For objective design, the analysis reframes log-sum-exp as a structural requirement rather than a numerical convenience. Softmax is often introduced to avoid overflow or to produce well-behaved gradients, but by the analysis of Section~\ref{sec:conditions}, LSE structure is also what induces the competition that produces responsibility-weighted learning. If inference-like behavior is desired, LSE structure is required; if independent predictions or robust outlier handling are preferred, it should be deliberately avoided. The choice determines what kind of learning dynamics the objective will produce.

For theory, the framework connects two processes traditionally kept distinct: optimization concerns finding parameters that minimize a loss, while inference concerns computing posteriors over latent variables. The implicit EM result shows that under distance-based LSE objectives, these are the same process viewed at different levels: the forward pass computes posteriors and the backward pass applies them. This is not a claim that all optimization is inference; for a well-defined class of objectives, however, the distinction disappears.

\subsection{Open Directions}

Several directions remain open. The absence of volume control in most neural objectives, the missing log-determinant, leads to collapse risks currently managed by heuristics. Section~\ref{sec:volume} supplies the exact term for the linear-layer case, where the volume correction is a log-determinant of the same weights that compute the distance. What remains open is the general case: deriving implicit volume terms for deep, nonlinear compositions of distance layers, and understanding when normalization layers substitute for explicit volume control and when they do not. Answering this would connect the framework to practical stability concerns across architectures.

Supervision in real settings is rarely clean. Labels may be noisy, partial, or uncertain. The constrained regime analysis assumes hard labels that clamp responsibilities exactly; a fuller treatment would model soft or probabilistic supervision as partial constraints on the responsibility structure. This could bring semi-supervised learning, label smoothing, and learning from crowds under the same analysis.

Open-set inference requires relaxing the closed-world assumption. Current objectives force every input to be assigned; realistic deployment requires the option to reject. Objectives that support non-assignment, whether through an explicit ``none of the above'' component or a threshold below which no component takes responsibility, would extend implicit EM to settings where not all inputs belong to known categories.

Finally, the framework is falsifiable. At the loss level, the identity predicts that responsibilities can be read directly from gradients; over training, it predicts responsibility-driven specialization and, absent volume control, rich-get-richer dynamics. Tools that extract responsibilities from gradients, track specialization over training, and detect degeneration would move the framework from explanatory theory to practical instrument.

%% file: sec-conclusion.tex
\section{Conclusion}

This paper establishes a single result: for objectives of the form $L = \log \sum_j \exp(-d_j)$, the gradient with respect to each distance is the negative responsibility of the corresponding component. The identity $\partial L / \partial d_j = -r_j$ is exact and requires only that the distances be differentiable.

The implication is immediate: gradient descent on distance-based log-sum-exp objectives performs a generalized expectation-maximization implicitly. Responsibilities appear as the gradients themselves rather than as auxiliary quantities, with the forward pass playing the role of the E-step and the backward pass the M-step. No explicit inference algorithm is required because inference is embedded in the optimization.

This mechanism unifies phenomena that have been treated as distinct. Unsupervised mixture learning, attention in transformers, and cross-entropy classification are three regimes of the same underlying process, differing in what is observed and what is latent but governed by identical dynamics. The responsibility-weighted gradient structure recently documented in transformers is a necessary consequence of the objectives used to train them, not a property of the architecture. Under these objectives, optimization and inference are the same process viewed at different levels.